\pdfoutput=1

\documentclass[11pt]{article}

\usepackage[]{acl}

\usepackage{times}
\usepackage{latexsym}
\usepackage{graphicx}
\usepackage{amssymb}

\usepackage[T1]{fontenc}

\usepackage[utf8]{inputenc}

\usepackage{microtype}

\usepackage{inconsolata}
\usepackage{multirow}
\usepackage{amsmath}

\title{Enhancing CLIP Conceptual Embedding through Knowledge Distillation}

\author{Kuei-Chun Kao \quad \\
  University of California, Los Angeles \\ 
  \texttt{johnson0213@g.ucla.edu} \\}

\begin{document}
\maketitle
\begin{abstract}
Recently, CLIP has become an important model for aligning images and text in multi-modal contexts. However, researchers have identified limitations in the ability of CLIP's text and image encoders to extract detailed knowledge from pairs of captions and images. In response, this paper presents Knowledge-CLIP, an innovative approach designed to improve CLIP's performance by integrating a new knowledge distillation (KD) method based on Llama 2. Our approach focuses on three key objectives: Text Embedding Distillation, Concept Learning, and Contrastive Learning. First, Text Embedding Distillation involves training the Knowledge-CLIP text encoder to mirror the teacher model, Llama 2. Next, Concept Learning assigns a soft concept label to each caption-image pair by employing offline K-means clustering on text data from Llama 2, enabling Knowledge-CLIP to learn from these soft concept labels. Lastly, Contrastive Learning aligns the text and image embeddings. Our experimental findings show that the proposed model improves the performance of both text and image encoders.
\end{abstract}

\section{Introduction}
 The pre-training of the multimodal encoders associated with vision language, a good example being CLIP~\cite{radford2021learning}, has been found to be very helpful in learning transferrable features derived from paired data of image and text. The CLIP learning framework is contrastive, typically relying on data augmentation in order to eliminate unnecessary insertion and shortcuts. 
 
 However, the impressive results demonstrated by Vision and Language models (VLMs)~\cite{gokhale2023benchmarking,thrush2022winoground, yuksekgonul2023visionlanguage} on myriad recognized benchmarks do not necessarily indicate a comprehensive understanding of the compositional elements of text or images. These models, as exemplified by CLIP, raise questions about their ability to differentiate between sentence structures like "an orangutan eating and an officer flying" and "an orangutan and an officer eating an orangutan" Scenes in nature pose a significant challenge due to their complexity, resulting from the numerous objects and attributions they carry and their mutual interactions. 

The challenge inherent in CLIP, stemming from its difficulty in addressing image segmentation and object detection tasks due to the need for per-pixel label knowledge, has been addressed by \citet{li2022grounded} through the introduction of GLIP. The proposed method unified object detection and phrase grounding through pre-training, effectively leveraging external knowledge, i.e., grounding boxes. The integration of such external information facilitates the alignment of image-language data, enhancing the model's capability in handling complex visual tasks.



Inspired by their work, our objective is to integrate external knowledge from established large language models into CLIP, with the aim of further elevating its overall quality. Therefore, we introduce a pioneering methodology named Knowledge-CLIP, comprising triple sets of objectives. Firstly, our focus is on knowledge distillation (KD) from large language models, exemplified by Llama 2~\cite{touvron2023llama}, with the aim of enhancing the quality of CLIP's text encoder. Secondly, we posit that the embeddings generated by Llama 2 encompass more valuable attributes and conceptual information, such as color and action, than CLIP's text encoder. Thus, employing K-means clustering~\cite{hartigan1979algorithm} on Llama 2's embeddings, we derive soft concept labels for caption-image pairs. Subsequently, we leverage these soft concept labels to refine the quality of both CLIP's text and image encoders. Finally, we reused the contrastive objective from CLIP to continually align the text and image embeddings.

\begin{figure*}[t]
\begin{center}
\includegraphics[width=0.9\textwidth]{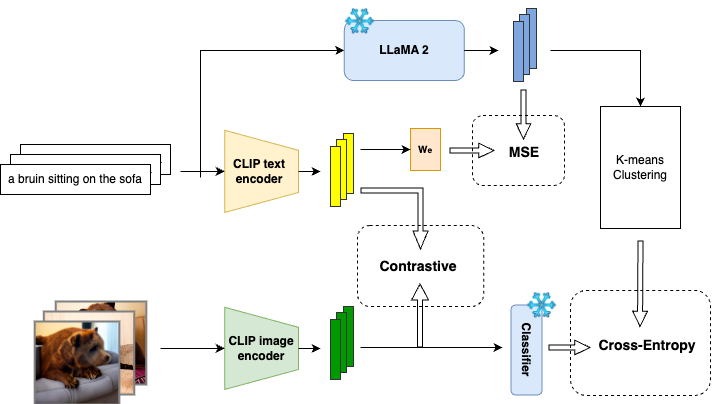}

\end{center}
   \caption{Overview of our proposed Knowledge-CLIP, which has five modules: CLIP text encoder ($E_{T}$), CLIP image encoder ($E_{I}$), Classifier ($C$), Linear projector ($W_e$) and Llama 2 ($L$).}
\label{fig:overview}
\end{figure*}
\section{Related Works}

\subsection{Knowledge Distillation}
Knowledge distillation (KD) is a training technique wherein a neural network, referred to as a student, is trained to replicate parts of another neural network, known as a teacher~\cite{ba2014deep}. The most common approach involves matching the output of the teacher network. However, an alternative option is to match the hidden layers, providing a more nuanced transfer of knowledge~\cite{romero2015fitnets, aguilar2020knowledge}. In terms of the loss functions utilized during this process, KL divergence is a common choice for matching probability outputs, while L2 norm is frequently employed to align the hidden vectors~\cite{kim2021comparing}. This technique allows for the compact representation of knowledge learned by the teacher network to be transferred to the student network, enhancing the student's performance and generalization capabilities.

\subsection{CLIP's Text Encoder}
In the context of multimodal models, the text-to-image query such as "A UCLA CS student in a futuristic lab, donned in virtual reality gear and programming a robotic assistant next to a professor in a lab coat." exemplifies the expectations for modern multimodal models. This type of query demands spatial precision (e.g., specifying the position of entities), compositional understanding (highlighting certain attributes like a UCLA CS student but not a UCLA CS assistant), and a touch of imagination, describing scenarios that may not exist in reality. However, recent works~\cite{gokhale2023benchmarking,thrush2022winoground, yuksekgonul2023visionlanguage} shed light on a notable challenge. Despite achieving robust benchmark performance, various multimodal models often struggle with even basic reasoning tasks, particularly those involving spatial relations or attribute attachments. These findings underscore the existing limitations in the reasoning capabilities of multimodal models, especially when confronted with intricate and imaginative textual input.

\section{Method}

\subsection{Problem Definition and Annotations}
For the sake of completeness, we first define the setting and notations considered in this paper. During training, we have $N$ caption-image pairs denoted as $X = \{X_{1}, X_{2}, ..., X_{N}\}$. For the $i$th caption-image pair, we have a caption and an image, i.e., $X_i = (x^{C}_i, x^{I}_i)$. As shown in Figure \ref{fig:overview}, our proposed Knowledge-CLIP has five modules: CLIP text encoder ($E_{T}$), CLIP image encoder ($E_{I}$), Classifier ($C$), Linear projector ($W_e$) and Llama 2 ($L$).

\subsection{Text Embedding Distillation}
In a previous KD work~\cite{jiao2019tinybert}, they effectively distill the knowledge from BERT$_{base}$ to TinyBERT with $mean$ $squared$ $error$ loss function (MSE). Thus we distill the output embeddings from Llama 2, and the objective is as follows:
\begin{equation}
\text{\normalsize}  L_{emb} = \sum_{i=1}^{N}\|(E_{T}(x_{i}^{C}))W_{e}-L(x_{i}^{C})\|_2^2\label{eq:1},
\end{equation}
where the matrix $W_e \in \mathbf{R} ^{d \times d'}$. The scalar values $d$ and $d'$ denote the hidden sizes of the outputs from CLIP text encoder ($E_T$) and Llama 2 ($L$). Noted that the matrix $W_e$ is a learnable linear transformation, which transforms the hidden states of $E_T$ into the same space as the $L$.

\subsection{Concept Learning}
Given a caption, there exist multiple attributes or concepts, e.g., color, position, action, etc. However, CLIP text encoder failed to extract these information observed from previous papers. We hypothesize that captions having the same attributes and concepts would have similar embeddings from Llama 2. Therefore, we utilize k-means clustering method to categorize the output embeddings from Llama 2 with inputting captions. 
Then, we regard the result from k-means clustering as the soft concept labels, denoted as $S = \{s_1, s_2, ..., s_N\}$. Noted that $s_i \in [K]$ is a K-class categorical variable.

After obtaining soft concept labels, we use the Llama 2's embeddings and their corresponding soft labels to train the Classifier ($C$). Later, we freeze the Classifier ($C$) and use it to train CLIP's image encoder with images and their corresponding soft labels. The objective is as follows:
\begin{equation}
\text{\normalsize}  L_{conc}=-\sum_{i=1}^{K} \sum_{j=1}^{N}s_{i} *log(C(E_I(x_{j}^I))),\label{eq:2}
\end{equation}
where $s_i$ denotes the ground truth one-hot vector representing the soft concept label.

\subsection{Contrastive Learning}
\begin{figure}[h]
\centering
\includegraphics[width=0.9\columnwidth]{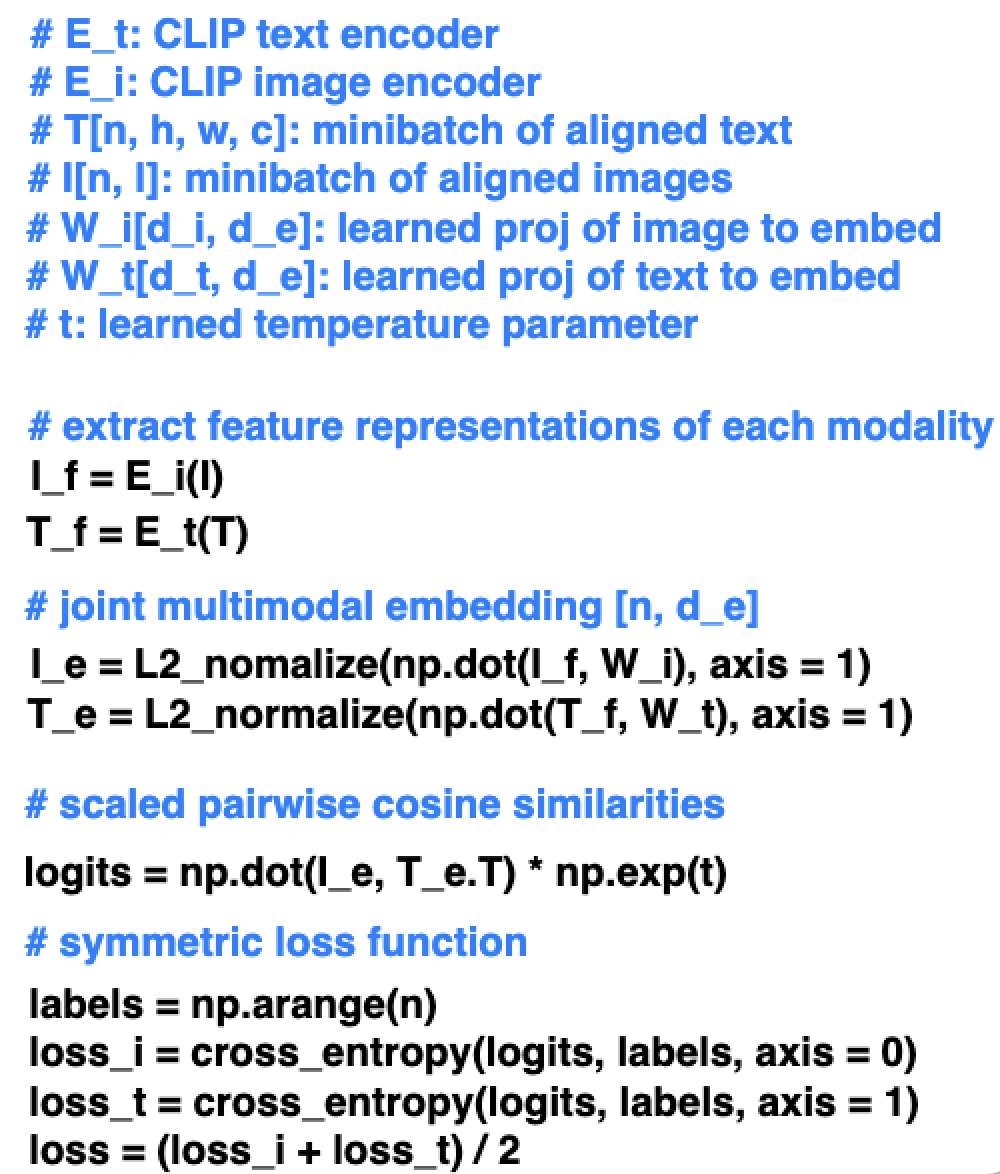}
\caption{Numpy-like pseudocode for the core of an implementation of CLIP.}
\label{fig:contrastive}
\end{figure}

In continuation of our exploration inspired by CLIP, our approach involves leveraging contrastive loss to effectively align text and image embeddings. Through this, we aim to optimize a symmetric cross-entropy loss, named $L_{cont}$, based on similarity scores. Figure~\ref{fig:contrastive} provides a visual representation of the pseudocode, outlining the core elements of an implementation of CLIP.
\subsection{Learning Objective}
We combine Text Embedding Distillation, Concept Learning, and Contrastive Learning to further improve the quality of CLIP encoders, i.e., text encoder, image encoder. The learning objective is as follows:
\begin{equation}
\text{\normalsize}  L = \alpha L_{emb} + \beta L_{conc} + \gamma L_{cont},
\label{eq:3}
\end{equation}
where $\alpha$, $\beta$, and $\gamma$ is a hyper parameter.

\begin{figure*}[t]
\begin{center}
\includegraphics[width=0.9\textwidth]{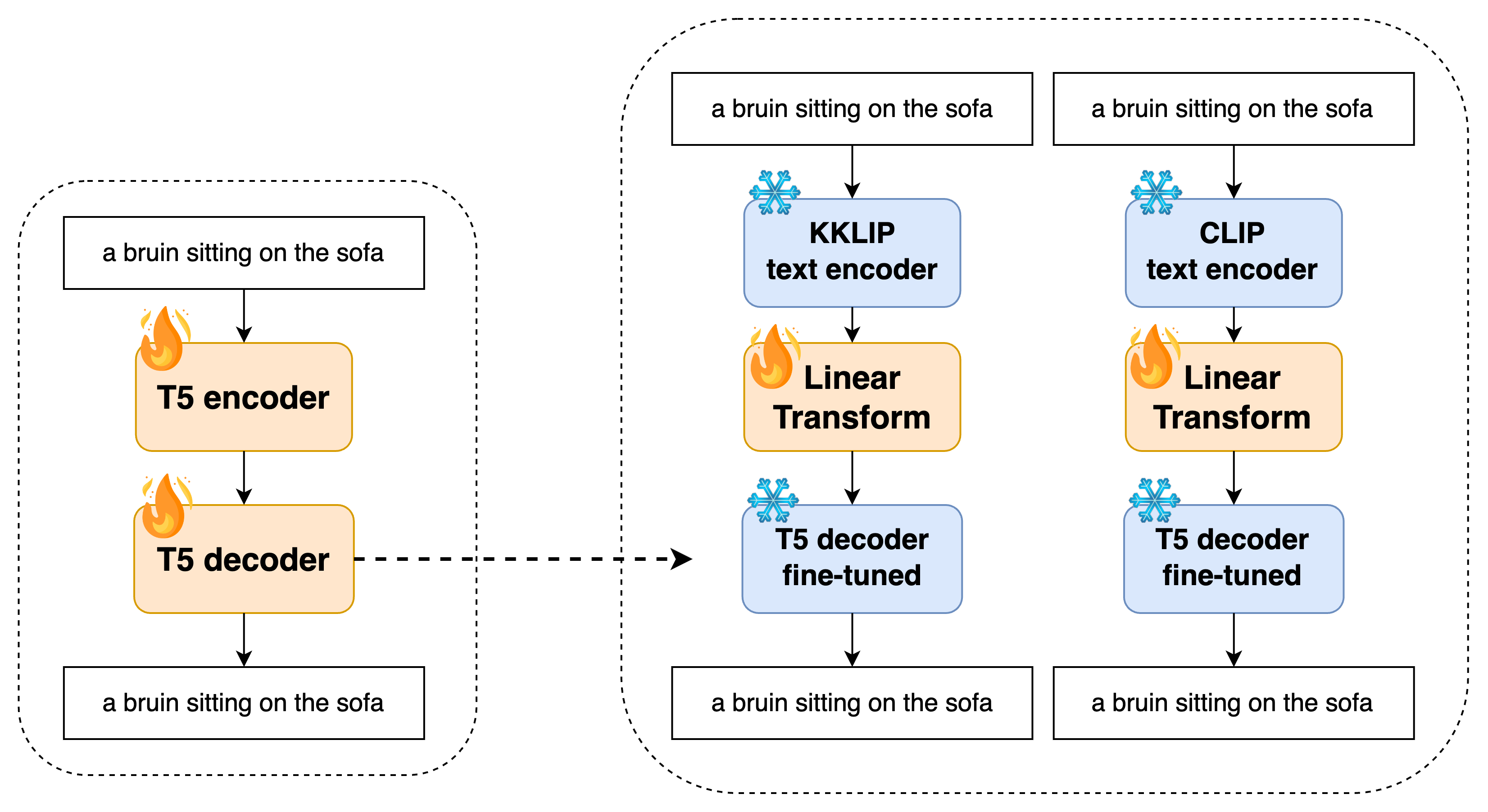}

\end{center}
   \caption{The evaluation process of text encoders}
\label{fig:text_encoder_test}
\end{figure*}
\section{Experiment}

\subsection{Experimental settings}

The pre-training dataset\footnote{\url{https://huggingface.co/datasets/yxchng/cc15m_yfcc15m}} comprises a subset of CLIP's pretraining data. With a total of 15.7 million records, we selectively extracted 500,000 records for model training and 100,000 records for model evaluation. We run 100 epoch and learning rate is $5 \times 10^{-5}$. $\alpha$ and $\gamma$ are equal to 1, and $\beta$ is equal to $0.01$.

\subsection{Knowledge-CLIP's Text Encoder}
\begin{table}[ht]
    \begin{tabular}{l|l}
        Encoder                          & EM (\%) \\ \hline
        Proof-of-concept T5              & 90     \\
        Llama2                           & 52.5   \\
        CLIP                             & 35.8   \\
        Knowledge-CLIP                            & 44.4  
    \end{tabular}
    \caption{EM of Different Text Encoder on CC3M.}
    \label{tab:text_encoder}
\end{table}

To assess the enhancement in Knowledge-CLIP's text encoder, we adopt the evaluation criteria outlined by Kamath et al.~\cite{kamath2023text}. The evaluation process, illustrated in Figure \ref{fig:text_encoder_test}, follows a multi-stage approach.

During the training stage, a T5-encoder is fine-tuned to generate text embeddings based on input sentences. Simultaneously, a T5-decoder is fine-tuned to produce output sentences that closely match the input sentences. Following this, the Knowledge-CLIP's and CLIP's text encoders, along with the T5-decoder, are frozen. The linear transformation layer and layer normalization are then fine-tuned to optimize their performance.

At the evaluation stage, the models are fixed, and output sentences are generated based on input sentences. The evaluation metric employed is Exact Match (EM), indicating the extent to which the generated output matches the input sentence verbatim.

Table \ref{tab:text_encoder} shows the exact match rate of various text encoders on the CC3M dataset. Initially, we present a proof-of-concept T5 Encoder to validate the feasibility of our experimental setup. Subsequently, the results reveal that Knowledge-CLIP's performance falls between that of Llama2 and CLIP. This observation suggests that the text embedding distillation employed in Knowledge-CLIP's text encoder enhances its performance, as the embeddings generated by Llama2 encapsulate a broader range of valuable attributes and conceptual information.

\subsection{Knowledge-CLIP's Image Encoder}
In this study, our primary objective is to assess the quality of Knowledge-CLIP's image encoder. 
While traditional zero-shot learning tasks such as classification typically overlook attribute values, assuming them to be understood, our approach involves a more detailed examination of these semantic features and their associated attributes (e.g., color, shape, leg, head, etc) which necessitate more sophisticated image encoder's features to align with the specific text.

Under this background, our evaluation experiment will first be assigned to a specific class for each image. Then, we predict the attribute descriptions based on their given class. This experiment explores the extent to which the image features generated by the image encoder can comprehend the conceptual meaning of kmeans soft labels, and how effectively our proposed models can bridge the gap between the learning of semantic entities and attribute-based recognition.

To address this question, we leverage two common attribute-based datasets, the AWA2~\cite{xian2020zeroshot} and CUB~\cite{Wah2011TheCB}, and explore the effect of attribute-based learning on our proposed model.

\textbf{\emph{CUB.}} The Caltech-UCSD Birds-200-2011~\cite{Wah2011TheCB} dataset features 11,788 images of 200 distinct bird species. Each species is annotated with 312 binary attributes. CUB provides attributes for each image, consisting of an attribute description and an expression. For instance, an attribute might have the description <attr>: "has head pattern" and the expression <expr>: "crested". Each attribute description has various possible expressions. CUB also provides class attributes, giving a probability that the attribute can be found in an image of that class for each class and attribute. For each attribute description, we select a maximum of one attribute with the highest probability. Therefore, the evaluation prompt would be "a photo of a <class label> that <attr> <expr>."

\textbf{\emph{AWA2.}} The Animals with Attributes 2 ~\cite{xian2020zeroshot} dataset includes 37,322 images of 50 animal classes. Each class is annotated with 85 binary attributes. Unlike CUB, AWA2's attributes do not have separate attribute descriptions and expressions. As such, we append the attributes as a comma-separated list at the end of the prompt. To maintain comparability with CUB, our evaluation prompt would be "a photo of a <class label>, with attribute <attr1>, <attr2>, ...".

During evaluation, for each image, we create $\mathbf{A}$ prompts for each class $\mathbf{C}$, where $\mathbf{C}$ is the number of classes and $\mathbf{A}$ is the number of attributes in dataset $\mathbf{D}$. We then calculate the cosine similarities between the image $\mathbf{I}$ and the $\mathbf{A}$ prompts and apply softmax to the similarity values. The attribute that corresponds to the prompt that is most similar to image $\mathbf{I}$ is the predicted class.


\begin{table}[ht]
\centering  
\begin{tabular}{|c|cc|cc|}
\hline
\multirow{2}{*}{Model/Dataset} & \multicolumn{2}{c|}{AWA2}          & \multicolumn{2}{c|}{CUB}           \\ \cline{2-5} 
                              & \multicolumn{1}{c|}{Top 1} & Top 5 & \multicolumn{1}{c|}{Top 1} & Top 5 \\ \hline
CLIP                           & \multicolumn{1}{c|}{55.8}  & 64.9  & \multicolumn{1}{c|}{78.3}  & 84.3  \\ \hline
Knowledge-CLIP                          & \multicolumn{1}{c|}{56.7}  & 65.1  & \multicolumn{1}{c|}{78.8}  & 85.0  \\ \hline
\end{tabular}
\caption{Results of using pre-trained CLIP and our proposed KKLIP model to evaluate AWA2 and CUB dataset.}
\label{tbl:model_performane}
\end{table}

The result in Table \ref{tbl:model_performane} shows that this setup is slightly beneficial for Knowledge-CLIP on AWA2 and also on CUB, in which cases our model is able to learn slightly better conceptual image features compared to the CLIP model, but also notice that this performance increase is not strong. 

\begin{figure*}[t]
\begin{center}
\includegraphics[width=0.9\textwidth]{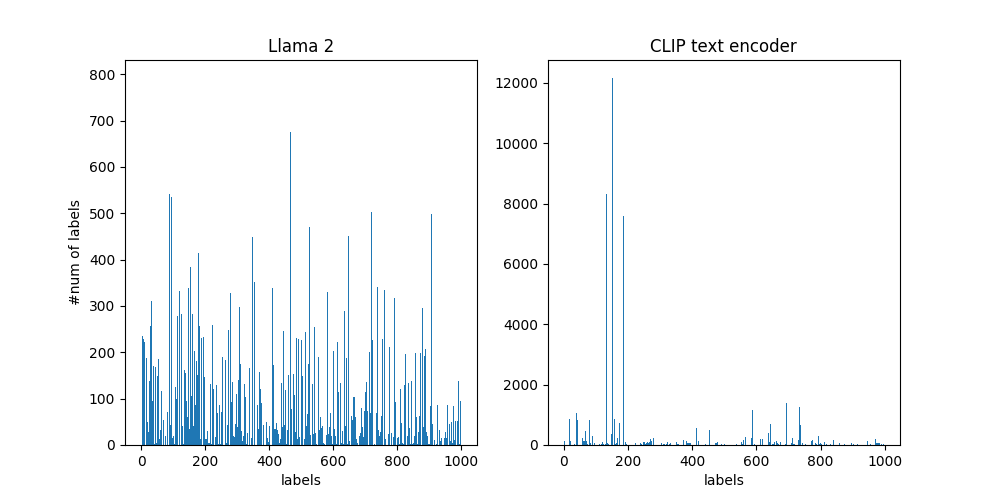}

\end{center}
   \caption{The distribution of text embedding generated by Llama 2 and CLIP}
\label{fig:explain}
\end{figure*}
\begin{figure}[ht]
\centering
\includegraphics[width=0.9\columnwidth]{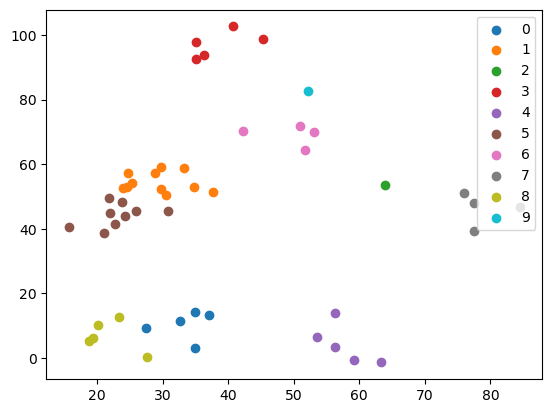}

\caption{Visualization of Llama 2's embeddings with different soft concept labels. Colors represent different attributes in CUB~\cite{Wah2011TheCB} dataset. }
\label{fig:kmeans_evaluation}
\end{figure}
\section{Discussion}

In this section, we delve into the distinctions between Llama 2 and CLIP's text encoder, shedding light on the unique attributes that set them apart. Additionally, we explore the potential significance of soft labels derived from k-means clustering applied to Llama 2's embeddings.
\subsection{Llama 2 v.s. CLIP}
To demonstrate the potential improvement in CLIP's text embeddings through the utilization of Llama 2, we propose an analysis of the distributional characteristics of their respective embeddings under identical dataset settings. Our hypothesis posits that Llama 2's text embeddings exhibit a more uniform distribution compared to those of CLIP since Llama 2 extracts more detailed text information than CLIP.

\textbf{Embedding Extraction:}
Text embeddings for the training dataset were extracted using both Llama 2 and CLIP text encoder. This initial step ensured that the embeddings were generated under identical conditions, facilitating a direct comparison.

\textbf{Clustering Analysis:}
To analyze the distribution of embeddings, we employed k-means clustering with 1000 clusters for each model. This unsupervised clustering method enabled us to discern patterns and groupings within the high-dimensional embedding space.

From figure \ref{fig:explain}, it is evident that the distribution of Llama 2's embeddings is notably more uniform compared to CLIP's. The visualization provides insights into the dispersion of data points in the embedding space, supporting our claim of enhanced uniformity in Llama 2's embeddings. Additionally, an examination of the most common labels within each clustering reveals a notable distinction. Llama 2 tends to have fewer occurrences of the most common labels, indicating a more diverse representation of information compared to CLIP.

This observed uniformity in Llama 2's embeddings suggests its potential to capture a broader range of semantic nuances, enabling finer distinctions between sentences. The reduced concentration on common labels further implies that Llama 2 may offer a richer representation that aids in discerning subtle differences between sentences.

Based on these findings, we propose leveraging Llama 2's embeddings for fine-tuning CLIP. The potential of Llama 2 to extract more nuanced information from textual data could enhance CLIP's ability to differentiate between sentences, contributing to improved overall performance in downstream tasks. Subsequent sections delve into the fine-tuning process and assess the impact on zero-shot classification accuracy.

\subsection{Visualization of K-means clustering with Llama 2}
To show what is learned from our proposed soft concpetual labels, we randomly select 50 samples from a specific class with different attributes from CUB dataset. (e.g. the following experiment uses "Black footed Albatross") Then, we cluster the llama2 embedding and visualize the according cluster.

In Figure \ref{fig:kmeans_evaluation}, it is clear that the llama2 embedding from each attribute is clustered into same groups, indicating that our method with K-means soft labels can better utilize conceptual learning.
\section{Conclusion}

In this paper, we introduce oduces Knowledge-CLIP, a novel methodology designed to enhance the overall quality of CLIP, a multimodal vision-language model. Our approach leverages a large language model, Llama 2, to guide both the image and text encoders. The experimental results demonstrate the effectiveness of Knowledge-CLIP in improving the quality of CLIP's text encoder and CLIP's image encoder. Through a comprehensive evaluation on the CC3M dataset, we observe that Knowledge-CLIP achieves a higher Exact Match rate compared to CLIP. Moreover, the evaluation of Knowledge-CLIP on image encoder quality, using attribute-based datasets AWA2 and CUB, indicates a slight performance improvement over CLIP. In conclusion, Knowledge-CLIP presents a promising approach for enhancing the capabilities of multimodal vision-language models like CLIP by incorporating external knowledge, refining embeddings, and addressing specific limitations. However, further investigation and experimentation may be necessary to optimize the model for specific tasks and domains. Future work could involve exploring additional datasets, refining hyperparameters, and investigating the model's applicability to different downstream tasks in vision-language understanding.

\bibliography{custom}

\begin{thebibliography}{15}
\expandafter\ifx\csname natexlab\endcsname\relax\def\natexlab#1{#1}\fi

\bibitem[{Aguilar et~al.(2020)Aguilar, Ling, Zhang, Yao, Fan, and Guo}]{aguilar2020knowledge}
Gustavo Aguilar, Yuan Ling, Yu~Zhang, Benjamin Yao, Xing Fan, and Chenlei Guo. 2020.
\newblock \href {http://arxiv.org/abs/1910.03723} {Knowledge distillation from internal representations}.

\bibitem[{Ba and Caruana(2014)}]{ba2014deep}
Lei~Jimmy Ba and Rich Caruana. 2014.
\newblock \href {http://arxiv.org/abs/1312.6184} {Do deep nets really need to be deep?}

\bibitem[{Gokhale et~al.(2023)Gokhale, Palangi, Nushi, Vineet, Horvitz, Kamar, Baral, and Yang}]{gokhale2023benchmarking}
Tejas Gokhale, Hamid Palangi, Besmira Nushi, Vibhav Vineet, Eric Horvitz, Ece Kamar, Chitta Baral, and Yezhou Yang. 2023.
\newblock \href {http://arxiv.org/abs/2212.10015} {Benchmarking spatial relationships in text-to-image generation}.

\bibitem[{Hartigan and Wong(1979)}]{hartigan1979algorithm}
John~A Hartigan and Manchek~A Wong. 1979.
\newblock Algorithm as 136: A k-means clustering algorithm.
\newblock \emph{Journal of the royal statistical society. series c (applied statistics)}, 28(1):100--108.

\bibitem[{Jiao et~al.(2019)Jiao, Yin, Shang, Jiang, Chen, Li, Wang, and Liu}]{jiao2019tinybert}
Xiaoqi Jiao, Yichun Yin, Lifeng Shang, Xin Jiang, Xiao Chen, Linlin Li, Fang Wang, and Qun Liu. 2019.
\newblock Tinybert: Distilling bert for natural language understanding.
\newblock \emph{arXiv preprint arXiv:1909.10351}.

\bibitem[{Kamath et~al.(2023)Kamath, Hessel, and Chang}]{kamath2023text}
Amita Kamath, Jack Hessel, and Kai-Wei Chang. 2023.
\newblock \href {http://arxiv.org/abs/2305.14897} {Text encoders bottleneck compositionality in contrastive vision-language models}.

\bibitem[{Kim et~al.(2021)Kim, Oh, Kim, Cho, and Yun}]{kim2021comparing}
Taehyeon Kim, Jaehoon Oh, NakYil Kim, Sangwook Cho, and Se-Young Yun. 2021.
\newblock \href {http://arxiv.org/abs/2105.08919} {Comparing kullback-leibler divergence and mean squared error loss in knowledge distillation}.

\bibitem[{Li et~al.(2022)Li, Zhang, Zhang, Yang, Li, Zhong, Wang, Yuan, Zhang, Hwang et~al.}]{li2022grounded}
Liunian~Harold Li, Pengchuan Zhang, Haotian Zhang, Jianwei Yang, Chunyuan Li, Yiwu Zhong, Lijuan Wang, Lu~Yuan, Lei Zhang, Jenq-Neng Hwang, et~al. 2022.
\newblock Grounded language-image pre-training.
\newblock In \emph{Proceedings of the IEEE/CVF Conference on Computer Vision and Pattern Recognition}, pages 10965--10975.

\bibitem[{Radford et~al.(2021)Radford, Kim, Hallacy, Ramesh, Goh, Agarwal, Sastry, Askell, Mishkin, Clark et~al.}]{radford2021learning}
Alec Radford, Jong~Wook Kim, Chris Hallacy, Aditya Ramesh, Gabriel Goh, Sandhini Agarwal, Girish Sastry, Amanda Askell, Pamela Mishkin, Jack Clark, et~al. 2021.
\newblock Learning transferable visual models from natural language supervision.
\newblock In \emph{International conference on machine learning}, pages 8748--8763. PMLR.

\bibitem[{Romero et~al.(2015)Romero, Ballas, Kahou, Chassang, Gatta, and Bengio}]{romero2015fitnets}
Adriana Romero, Nicolas Ballas, Samira~Ebrahimi Kahou, Antoine Chassang, Carlo Gatta, and Yoshua Bengio. 2015.
\newblock \href {http://arxiv.org/abs/1412.6550} {Fitnets: Hints for thin deep nets}.

\bibitem[{Thrush et~al.(2022)Thrush, Jiang, Bartolo, Singh, Williams, Kiela, and Ross}]{thrush2022winoground}
Tristan Thrush, Ryan Jiang, Max Bartolo, Amanpreet Singh, Adina Williams, Douwe Kiela, and Candace Ross. 2022.
\newblock \href {http://arxiv.org/abs/2204.03162} {Winoground: Probing vision and language models for visio-linguistic compositionality}.

\bibitem[{Touvron et~al.(2023)Touvron, Martin, Stone, Albert, Almahairi, Babaei, Bashlykov, Batra, Bhargava, Bhosale et~al.}]{touvron2023llama}
Hugo Touvron, Louis Martin, Kevin Stone, Peter Albert, Amjad Almahairi, Yasmine Babaei, Nikolay Bashlykov, Soumya Batra, Prajjwal Bhargava, Shruti Bhosale, et~al. 2023.
\newblock Llama 2: Open foundation and fine-tuned chat models.
\newblock \emph{arXiv preprint arXiv:2307.09288}.

\bibitem[{Wah et~al.(2011)Wah, Branson, Welinder, Perona, and Belongie}]{Wah2011TheCB}
Catherine Wah, Steve Branson, Peter Welinder, Pietro Perona, and Serge~J. Belongie. 2011.
\newblock \href {https://api.semanticscholar.org/CorpusID:16119123} {The caltech-ucsd birds-200-2011 dataset}.

\bibitem[{Xian et~al.(2020)Xian, Lampert, Schiele, and Akata}]{xian2020zeroshot}
Yongqin Xian, Christoph~H. Lampert, Bernt Schiele, and Zeynep Akata. 2020.
\newblock \href {http://arxiv.org/abs/1707.00600} {Zero-shot learning -- a comprehensive evaluation of the good, the bad and the ugly}.

\bibitem[{Yuksekgonul et~al.(2023)Yuksekgonul, Bianchi, Kalluri, Jurafsky, and Zou}]{yuksekgonul2023visionlanguage}
Mert Yuksekgonul, Federico Bianchi, Pratyusha Kalluri, Dan Jurafsky, and James Zou. 2023.
\newblock \href {http://arxiv.org/abs/2210.01936} {When and why vision-language models behave like bags-of-words, and what to do about it?}

\end{thebibliography}
\end{document}